\documentclass[10pt,twocolumn,letterpaper]{article}

\usepackage[pagenumbers]{cvpr} 

\usepackage{booktabs}
\usepackage{colortbl}
\usepackage{graphicx}
\usepackage{lipsum}
\usepackage{listings}
\usepackage{makecell}
\usepackage{mathtools}
\usepackage{multirow}
\usepackage{nicematrix}
\usepackage{float}
\usepackage{ifthen}

\let\svthefootnote\thefootnote
\newcommand\freefootnote[1]{%
\let\thefootnote\relax%
\footnotetext{#1}%
\let\thefootnote\svthefootnote%
}

\definecolor{t_red}{HTML}{FF9999}
\definecolor{t_orange}{HTML}{FFCC99}
\definecolor{t_yellow}{HTML}{FFF6B2}

\definecolor{cvprblue}{rgb}{0.21,0.49,0.74}
\usepackage[pagebackref,breaklinks,colorlinks,allcolors=cvprblue]{hyperref}

\def\paperID{884} 
\def\confName{CVPR}
\def\confYear{2025}

\title{
Beyond Gaussians: Fast and High-Fidelity 3D Splatting with Linear Kernels
}

\author{
Haodong Chen\textsuperscript{1} \quad
Runnan Chen\textsuperscript{1} \quad
Qiang Qu\textsuperscript{1} \quad
Zhaoqing Wang\textsuperscript{1} \\
Tongliang Liu\textsuperscript{1}\thanks{Hello} \quad
Xiaoming Chen\textsuperscript{2}\footnotemark[1] \quad
Yuk Ying Chung\textsuperscript{1}\footnotemark[1] \\
\textsuperscript{1}University of Sydney \quad
\textsuperscript{2} Beijing Technology and Business University\\
{\tt\small tongliang.liu@sydney.edu.au},\quad{\tt\small xiaoming.chen@btbu.edu.cn},\quad{\tt\small vera.chung@sydney.edu.au}
}

\begin{document}


\newlength{\ogheavyrulewidth}
\newlength{\oglightrulewidth}
\newlength{\ogcmidrulewidth}
\newlength{\ogaboverulesep}
\newlength{\ogbelowrulesep}

\setlength{\ogheavyrulewidth}{\heavyrulewidth}
\setlength{\oglightrulewidth}{\lightrulewidth}
\setlength{\ogcmidrulewidth}{\cmidrulewidth}
\setlength{\ogaboverulesep}{\aboverulesep}
\setlength{\ogbelowrulesep}{\belowrulesep}

\setlength{\heavyrulewidth}{0.07em}
\setlength{\lightrulewidth}{0.02em}
\setlength{\cmidrulewidth}{0.02em}
\setlength{\aboverulesep}{0.05em}
\setlength{\belowrulesep}{0.05em}
\renewcommand{\arraystretch}{0.6}

\newsavebox{\tableQuantitative}
\begin{lrbox}{\tableQuantitative}
\scriptsize
\begin{NiceTabular}{l|ccc|ccc|ccc}[cell-space-limits=0.1em,colortbl-like]
\toprule
\Block{2-1}{Method} & \multicolumn{3}{c}{Mip-NeRF360} & \multicolumn{3}{c}{Tanks\&Temples} & \multicolumn{3}{c}{Deep Blending} \\
 & SSIM $\uparrow$ & PSNR $\uparrow$ & LPIPS $\downarrow$ & SSIM $\uparrow$ & PSNR $\uparrow$ & LPIPS $\downarrow$ & SSIM $\uparrow$ & PSNR $\uparrow$ & LPIPS $\downarrow$ \\ \midrule
Plenoxels\textsuperscript{\textdagger}~\cite{FridovichKeil2022} & 0.626 & 23.08 & 0.463 & 0.719 & 21.08 & 0.379 & 0.795 & 23.06 & 0.510 \\
INGP-Base\textsuperscript{\textdagger}~\cite{Mueller2022} & 0.671 & 25.30 & 0.371 & 0.723 & 21.72 & 0.330 & 0.797 & 23.62 & 0.423 \\
INGP-Big\textsuperscript{\textdagger}~\cite{Mueller2022} & 0.699 & 25.59 & 0.331 & 0.745 & 21.92 & 0.305 & 0.817 & 24.96 & 0.390 \\
Mip-NeRF360\textsuperscript{\textdagger}~\cite{Barron2022} & 0.792 & \cellcolor{t_red} 27.69 & 0.237 & 0.759 & 22.22 & 0.257 & 0.901 & 29.40 & 0.245 \\
3DGS-7K\textsuperscript{\textdagger}~\cite{Kerbl2023} & 0.770 & 25.60 & 0.279 & 0.767 & 21.20 & 0.280 & 0.875 & 27.78 & 0.317 \\
3DGS-30K\textsuperscript{\textdagger}~\cite{Kerbl2023} & \cellcolor{t_yellow} 0.815 & 27.21 & \cellcolor{t_yellow} 0.214 & 0.841 & 23.14 & 0.183 & 0.903 & 29.41 & \cellcolor{t_orange} 0.243 \\
\midrule
3DGS~\cite{Kerbl2023} & \cellcolor{t_yellow} 0.815 & 27.47 & 0.216 & 0.847 & 23.62 & 0.178 & \cellcolor{t_orange} 0.905 & 29.49 & 0.247 \\
2DGS~\cite{Huang2024} & 0.805 & 27.20 & 0.231 & 0.830 & 22.89 & 0.203 & \cellcolor{t_yellow} 0.904 & \cellcolor{t_red} 29.67 & 0.249 \\
Mip-Splatting~\cite{Yu2024} & \cellcolor{t_yellow} 0.815 & 27.51 & 0.220 & 0.848 & \cellcolor{t_yellow} 23.65 & 0.181 & \cellcolor{t_red} 0.906 & \cellcolor{t_orange} 29.66 & 0.245 \\
AbsGS~\cite{Ye2024a} & \cellcolor{t_orange} 0.822 & 27.41 & \cellcolor{t_red} 0.186 & \cellcolor{t_yellow} 0.854 & 23.43 & \cellcolor{t_orange} 0.158 & 0.898 & 28.77 & 0.249 \\
\midrule
\textbf{3DLS (Ours)} & \cellcolor{t_orange} 0.822 & \cellcolor{t_yellow} 27.57 & \cellcolor{t_orange} 0.196 & \cellcolor{t_orange} 0.855 & \cellcolor{t_orange} 23.71 & \cellcolor{t_yellow} 0.160 & 0.902 & 29.44 & \cellcolor{t_yellow} 0.244 \\
\textbf{3DLS+AA (Ours)} & \cellcolor{t_red} 0.823 & \cellcolor{t_orange} 27.63 & \cellcolor{t_orange} 0.196 & \cellcolor{t_red} 0.860 & \cellcolor{t_red} 23.89 & \cellcolor{t_red} 0.157 & \cellcolor{t_yellow} 0.904 & \cellcolor{t_yellow} 29.50 & \cellcolor{t_red} 0.240 \\
\bottomrule
\end{NiceTabular}
\end{lrbox}

\setlength{\heavyrulewidth}{\ogheavyrulewidth}
\setlength{\lightrulewidth}{\oglightrulewidth}
\setlength{\cmidrulewidth}{\ogcmidrulewidth}
\setlength{\aboverulesep}{\ogaboverulesep}
\setlength{\belowrulesep}{\ogbelowrulesep}
\renewcommand{\arraystretch}{1}

\renewcommand{\arraystretch}{0.9}
\newsavebox{\tableAblation}
\begin{lrbox}{\tableAblation}
\begin{NiceTabular}{ccc|cccc|cccc|cccc}[colortbl-like]
\toprule
\multicolumn{3}{c}{Methods} & \multicolumn{4}{c}{Kitchen-30K} & \multicolumn{4}{c}{Train-30K} & \multicolumn{4}{c}{MEAN} \\
\textit{LK} & \textit{DA} & \textit{AGS} & SSIM $\uparrow$ & PSNR $\uparrow$ & LPIPS $\downarrow$ & N $\downarrow$ & SSIM $\uparrow$ & PSNR $\uparrow$ & LPIPS $\downarrow$ & N $\downarrow$ & SSIM $\uparrow$ & PSNR $\uparrow$ & LPIPS $\downarrow$ & N $\downarrow$ \\
\midrule
&  &  &
0.928 & 31.40 & 0.127 & 1,778,626 &
0.814 & 21.88 & 0.207 & 1,111,908 &
0.871 & \cellcolor{t_orange} 26.64 & 0.167 & \cellcolor{t_orange} 1,445,267 \\
\checkmark &  &  &
0.928 & 31.18 & 0.131 & 1,117,539 &
0.818 & 21.62 & 0.198 & 1,015,078 &
\cellcolor{t_yellow} 0.873 & 26.40 & \cellcolor{t_yellow} 0.165 & \cellcolor{t_red} 1,066,309 \\
\checkmark & \checkmark &  & 
0.932 & 31.75 & 0.118 & 3,269,909 &
0.826 & 21.51 & 0.178 & 3,199,961 &
\cellcolor{t_red} 0.879 & \cellcolor{t_yellow} 26.63 & \cellcolor{t_red} 0.148 & 3,234,935 \\
\checkmark & \checkmark & \checkmark & 
0.931 & 31.78 & 0.120 & 1,849,688 &
0.825 & 21.93 & 0.183 & 1,657,427 &
\cellcolor{t_orange} 0.878 & \cellcolor{t_red} 26.86 & \cellcolor{t_orange} 0.152 & \cellcolor{t_yellow} 1,753,558 \\
\bottomrule
\end{NiceTabular}
\end{lrbox}
\renewcommand{\arraystretch}{1}

\newsavebox{\tablePerformance}
\begin{lrbox}{\tablePerformance}
\begin{NiceTabular}{@{}l|cc|cc|cc|cc}
\toprule
\multirow{2}{*}{Method} &
\multicolumn{2}{c}{FPS [fwd]} & \multicolumn{2}{c}{FPS [bwd]} &
\multicolumn{2}{c}{Mem (MB)} & \multicolumn{2}{c}{N} \\ 
 & mean & $\Delta\%$ & mean & $\Delta\%$ & mean & $\Delta\%$ & mean & $\Delta\%$ \\ \midrule
3DGS  & 194.67 & --    & 112.78 & --    & 626.51 & --    & 2,655,026 & --    \\
Mip-Sp & 190.99 & -1.89 & 113.21 & +0.38  & 601.83 & -3.94 & 2,794,224 & +5.24  \\
AbsGS & 155.18 & -20.28 & 88.82 & -21.25 & 885.42 & +41.33 & 3,254,924 & +22.59 \\
\midrule
\textbf{3DLS}  & 260.83 & +33.99 & 148.65 & +31.80 & 658.92 & +5.17 & 3,349,999 & +26.18 \\
\bottomrule
\end{NiceTabular}
\end{lrbox}

\newsavebox{\tableParams}
\begin{lrbox}{\tableParams}
\begin{NiceTabular}{@{}c|cccccc@{}}
\toprule
\multirow{2}{*}{Method} & \multicolumn{6}{c}{Densification Parameters} \\
& $\tau_{G\nabla}$ & $\tau_{G2D}$ & $\tau_{G3D}$ & $\tau_{P2D}$ & $\tau_{P3D}$ & $\tau_{PO}$ \\
\midrule
3DGS & 0.0002 & 0.05 & 0.01 & 0.15 & 0.1 & 0.005 \\
\midrule
3DLS & 0.0002 & 0.05 & 0.006 & 0.15 & 0.4 & 0.025 \\
\bottomrule
\end{NiceTabular}
\end{lrbox}


\twocolumn[{
\renewcommand\twocolumn[1][]{#1}
\maketitle
\begin{center}
\centering
\captionsetup{type=figure}
\vspace{-3.5em}
\url{https://hche8927.github.io/3DLS/}
\includegraphics[width=\linewidth]{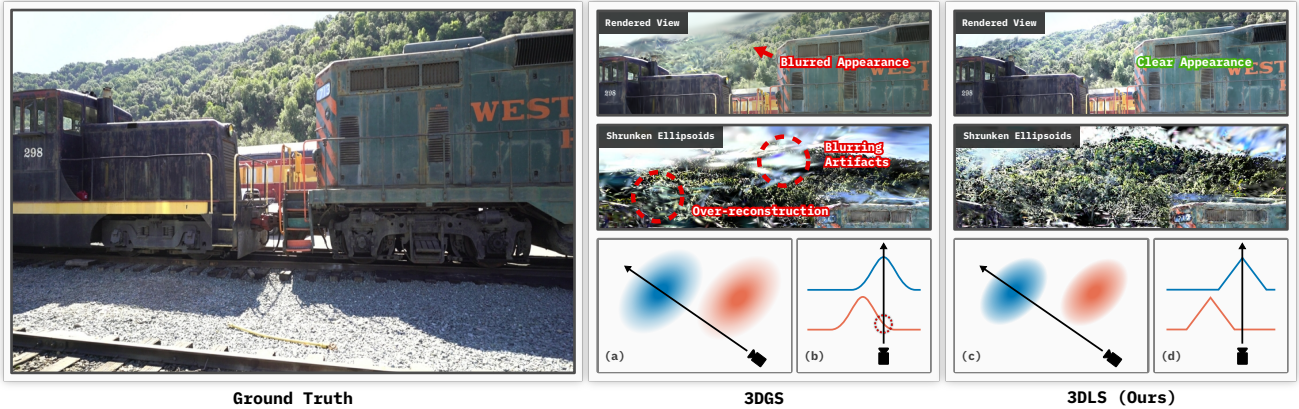}
\vspace{-1.8em}
\captionof{figure}{Comparison of 3D splatting with Gaussian and linear kernels. Gaussian kernel-based splatting results in blurred effects, floating artifacts, and over-reconstruction, where small-scale geometry is represented by oversized splats, reducing clarity in high-frequency regions. Panel (a) shows 3D Gaussian Splatting (3DGS)~\cite{Kerbl2023}, where soft ellipsoid boundaries cause interference between foreground and background. Panel (b) illustrates how the unbounded support of Gaussian kernels hinders separation in 1D distributions. In contrast, panels (c) and (d) show our 3D Linear Splatting (3DLS), where bounded linear kernels reduce interference and enhance separation, achieving clearer and more accurate reconstructions.}
\vspace{-0.8em}
\label{fig:title}
\end{center}
}]


\begin{abstract}
\label{section:abstract}
Recent advancements in 3D Gaussian Splatting (3DGS) have substantially improved novel view synthesis, enabling high-quality reconstruction and real-time rendering. However, blurring artifacts, such as floating primitives and over-reconstruction, remain challenging. Current methods address these issues by refining scene structure, enhancing geometric representations, addressing blur in training images, improving rendering consistency, and optimizing density control, yet the role of kernel design remains underexplored. We identify the soft boundaries of Gaussian ellipsoids as one of the causes of these artifacts, limiting detail capture in high-frequency regions. To bridge this gap, we introduce 3D Linear Splatting (3DLS), which replaces Gaussian kernels with linear kernels to achieve sharper and more precise results, particularly in high-frequency regions. Through evaluations on three datasets, 3DLS demonstrates state-of-the-art fidelity and accuracy, along with a 30\% FPS improvement over baseline 3DGS. The implementation will be made publicly available upon acceptance.
\freefootnote{*Corresponding author.}
\end{abstract}
\vspace{-2em}


\section{Introduction}
\label{section:introduction}

Rendering high-quality 3D content remains a core challenge in computer vision, with applications spanning neural rendering, virtual reality (VR), autonomous driving, and real-time simulation. Among recent advancements, 3D Gaussian Splatting (3DGS)~\cite{Kerbl2023} has gained prominence as an efficient, point-based approach to 3D rendering, utilizing continuous splats to compactly represent scenes. Despite the successes achieved by 3DGS, it encounters limitations in high-frequency regions with intricate textures and fine details, where artifacts such as blurring and floating primitives degrade the rendering quality. To address these issues, a range of methods has been developed to refine scene structure~\cite{Lu2024,Ververas2024}, enhance geometric representations~\cite{Huang2024,Waczynska2024,Gao2024,Guedon2024}, improve the handling of blurred training images~\cite{Lee2024,Zhao2024,Peng2024,Seiskari2024}, maintain rendering consistency~\cite{Yu2024,Yan2024}, and optimize density control~\cite{Ye2024a,Zhang2024}. While these approaches have led to improved detail capture and visual quality, artifacts remain persistent, especially in regions that require high-frequency detail and sharp transitions.

In this paper, we approach these limitations by examining kernel design in 3DGS. Our analysis reveals that Gaussian kernels are one of the causes of persistent artifacts, which can hinder the method's effectiveness. As shown in Figure \ref{fig:title}, Gaussian kernels produce ellipsoids with soft boundaries that complicate the separation of foreground and background primitives, resulting in artifacts such as floating primitives and over-smoothing. This blending across neighboring splats leads to ambiguous floating primitives that obscure sharp transitions and limit 3DGS's ability to accurately capture high-frequency details.

To tackle these issues, we propose \textit{3D Linear Splatting (3DLS)}, which replaces Gaussian kernels with linear kernels to improve the capture of high-frequency details. The bounded nature of linear kernels, in contrast to Gaussian kernels, minimizes interference among neighboring primitives, allowing for sharper transitions and more precise reconstructions. To further enhance 3DLS, we introduce two complementary techniques: \textit{Distribution Alignment (DA)} and \textit{Adaptive Gradient Scaling (AGS)}. Transitioning from one kernel distribution to another introduces differences in base spread that can disrupt distribution coverage. \textit{DA} addresses this challenge by aligning the spread of the linear kernel with Gaussian-based methods, ensuring compatibility with existing 3DGS frameworks and enhancing reconstruction fidelity. Additionally, changing kernel functions alters gradient calculations, impacting training stability. \textit{AGS} mitigates this by balancing detail preservation with computational efficiency, stabilizing training, and enabling 3DLS to effectively capture fine details, sharp transitions, and high-frequency content.

We validate 3DLS on three benchmark datasets, where it demonstrates state-of-the-art (SOTA) performance in both visual fidelity and accuracy. Additionally, 3DLS achieves a 30\% FPS increase over 3DGS with minimal memory overhead, making it highly suitable for real-time applications such as interactive rendering and VR.

In summary, our main contributions are:
\begin{enumerate}
    \item Introducing \textit{3D Linear Splatting (3DLS)}, a novel approach that replaces Gaussian kernels with linear kernels to improve rendering quality in high-frequency regions and offer a new perspective on kernel functions in splatting-based rendering.
    \item Proposing \textit{Distribution Alignment (DA)} to improve 3DLS integration with existing frameworks by aligning the kernel spread with Gaussian kernels.
    \item Proposing \textit{Adaptive Gradient Scaling (AGS)} to enhance 3DLS training stability and balance detail preservation with computational efficiency.
    \item Conducting extensive experiments on benchmark datasets, demonstrating both qualitative and quantitative improvements, including a 30\% increase in FPS.
\end{enumerate}


\section{Related Work}
\label{section:related_work}

This section reviews recent advancements in 3DGS and related rendering techniques, then focuses on methods aimed at enhancing the visual fidelity of 3DGS. For a comprehensive overview, we refer readers to recent excellent surveys~\cite{Dalal2024,Fei2024,Chen2024a,Wu2024}, which provide an in-depth analysis of the current progress and challenges in the field.

\subsection{Advances in 3D Gaussian-based Rendering}

3DGS~\cite{Kerbl2023} builds on earlier neural rendering methods~\cite{Penner2017,Henzler2019,Sitzmann2019,Mildenhall2021,Barron2022,Mueller2022,FridovichKeil2022} and point-based radiance fields~\cite{Lassner2021,Kopanas2022,Xu2022}, providing an efficient 3D scene representation through explicit point-based splats. Unlike volumetric approaches such as NeRF~\cite{Mildenhall2021} and Mip-NeRF~\cite{Barron2022}, which encode scenes within neural networks, 3DGS assigns Gaussian kernels to each point to directly model spatial extent and blending properties. This design enables smooth, continuous surface rendering with view-dependent effects like color and transparency. By adopting an explicit representation instead of a neural radiance field, 3DGS achieves high-speed, real-time rendering with minimal memory overhead, making it particularly suited for applications in autonomous driving~\cite{Zhou2024,Chen2024b,Yan2024b}, SLAM~\cite{Yan2024a,Huang2024a,Keetha2024,Matsuki2024,Ha2025,Li2025}, interactive simulations~\cite{Xiang2024,Jiang2024,Liu2024,Hu2024,Cao2024}, and content creation~\cite{Ren2024a,Ren2024b,Tang2024,Tang2025,Xie2024,Gao2024a}.

Despite its strengths, 3DGS faces challenges in high-frequency regions, often resulting in artifacts and blurry reconstructions. These limitations highlight the need for refined splatting techniques that better preserve detail while remaining computationally efficient.

\begin{figure*}[ht]
\centering
\includegraphics[width=\linewidth]{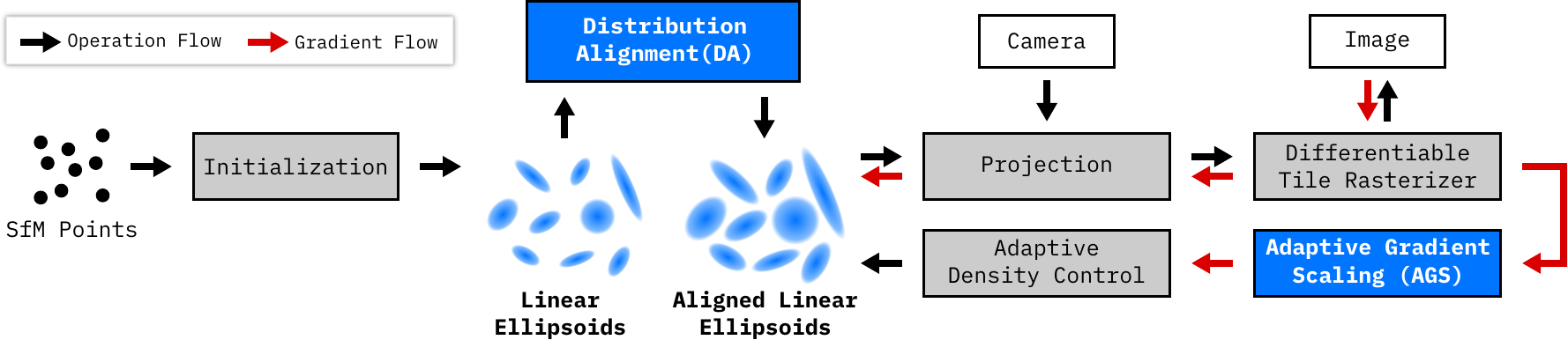}
\caption{Overview of our method integrated within the 3DGS framework. The process begins with replacing Gaussian kernels with linear kernels to enhance detail capture. Next, \textit{DA} ensures comprehensive splat coverage, optimizing compatibility with existing frameworks. Finally, \textit{AGS} is applied to support stable training and improve convergence, resulting in higher visual fidelity.}
\vspace{-1em}
\label{fig:flowchart}
\end{figure*}

\subsection{Toward High-Fidelity 3DGS}  
\label{section:toward_high_fidelity_3DGS}

Efforts to enhance 3DGS fidelity focus on five key areas: scene structure, geometric representation, handling blur in training images, rendering consistency, and density control. Techniques like Scaffold-GS~\cite{Lu2024} and SAGS~\cite{Ververas2024} improve scene structure through dynamic placement and graph-based optimization. Geometric representation is refined by using 2D disks~\cite{Huang2024} or extracting mesh-based structures~\cite{Waczynska2024,Gao2024,Guedon2024}. Handling blur in training images typically relies on neural networks~\cite{Lee2024,Zhao2024,Peng2024} or visual-inertial odometry~\cite{Seiskari2024}. Rendering consistency across resolutions is enhanced through 3D smoothing, Mip filters~\cite{Yu2024}, and multi-scale Gaussian splatting~\cite{Yan2024}. Density control techniques, like AbsGS~\cite{Ye2024a} and Pixel-GS~\cite{Zhang2024}, further improve detail capture with minimal memory overhead.

These advancements have further refined the fidelity and performance of 3DGS, extending its applicability across diverse scenarios. However, research on kernel design remains limited. Our proposed 3DLS addresses this gap by rethinking kernel design, replacing Gaussian kernels with linear ones to enhance high-frequency detail capture and accelerate rendering speed.


\section{Methods}
\label{section:methods}

While 3DGS~\cite{Kerbl2023} is effective at representing continuous, smooth surfaces, it encounters challenges when capturing high-frequency regions, such as fine details and intricate textures. These limitations stem from the inherent smoothness of the Gaussian kernel, which can introduce blurring and floating artifacts, especially in scenes with complex details. To overcome these challenges, we propose \textit{3D Linear Splatting (3DLS)}, which replaces Gaussian kernels with linear kernels to capture high-frequency details more effectively and enhance rendering clarity.

Figure~\ref{fig:flowchart} illustrates how our proposed method integrates within the existing 3DGS framework. First, Gaussian kernels are replaced with linear kernels to improve detail capture, as explained in Section~\ref{section:linear_kernel}. Next, \textit{Distribution Alignment (DA)} is introduced to ensure comprehensive splat coverage and compatibility with existing frameworks, described further in Section~\ref{section:distribution_alignment}. Finally, \textit{Adaptive Gradient Scaling (AGS)} is applied to support stable training and improve convergence, achieving higher visual fidelity and efficiency, as detailed in Section~\ref{section:adaptive_gradient_scaling}.

\subsection{Preliminary}

In 3DGS, each 3D primitive is modeled as a Gaussian distribution, represented by the following equation:
\begin{gather}
    G(\mathbf{x}) = \exp \left( -\frac{1}{2} (\mathbf{x} - \boldsymbol{\mu})^\top \Sigma^{-1} (\mathbf{x} - \boldsymbol{\mu}) \right), \\
    \Sigma = \mathbf{R} \mathbf{S} \mathbf{S}^\top \mathbf{R}^\top,
\end{gather}
where \(\mathbf{x} \in \mathbb{R}^3\) is a 3D point, \(\boldsymbol{\mu} \in \mathbb{R}^3\) represents the center of the Gaussian ellipsoid, and \(\Sigma\) denotes the covariance matrix. The covariance matrix \(\Sigma\) defines the shape, orientation, and spread of the Gaussian ellipsoid, with \(\mathbf{S}\) and \(\mathbf{R}\) representing the scaling and rotation matrices, respectively.

To project the 3D Gaussian ellipsoid onto the 2D image plane, the covariance matrix \(\Sigma\) is transformed into camera coordinates using the world-to-camera matrix \(\mathbf{W}\) and the local affine Jacobian matrix \(\mathbf{J}\):
\begin{equation}
    \Sigma' = \mathbf{J} \mathbf{W} \Sigma \mathbf{W}^\top \mathbf{J}^\top.
\end{equation}
The projected 2D covariance matrix \(\Sigma_{2D}\) is obtained by removing the third row and column of \(\Sigma'\), which correspond to the z-axis:
\begin{equation}
    \Sigma_{2D} = 
    \begin{bmatrix} 
        \Sigma'_{11} & \Sigma'_{12} \\
        \Sigma'_{21} & \Sigma'_{22}
    \end{bmatrix}.
\end{equation}

The resulting 2D Gaussian distribution on the image plane is then defined as:
\begin{equation}
    G'(\mathbf{x}') = \exp \left( -\frac{1}{2} (\mathbf{x}' - \boldsymbol{\mu}')^\top \Sigma_{2D}^{-1} (\mathbf{x}' - \boldsymbol{\mu}') \right),
\end{equation}
where \(\mathbf{x}' \in \mathbb{R}^2\) and \(\boldsymbol{\mu}' \in \mathbb{R}^2\) represent the projected 2D point and mean of the Gaussian, respectively.

To compute the pixel color at position \(\mathbf{x}'\), 3DGS applies alpha blending, accumulating contributions from multiple Gaussians along the corresponding ray in a front-to-back order:
\begin{equation}
    C(\mathbf{x}') = \sum_{i \in N} c_i \alpha_i \prod_{j=1}^{i-1} (1 - \alpha_j), \quad \alpha_i = o_i \cdot G'_i(\mathbf{x}').
\end{equation}
Here, \(c_i\) is the color of the \(i\)-th Gaussian, \(o_i\) denotes its opacity, and \(G'_i(\mathbf{x}')\) is the value of the 2D Gaussian at pixel position \(\mathbf{x}'\).

\subsection{Transition to Linear Kernels}
\label{section:linear_kernel}

Building on the foundational formulation of 3DGS, we introduce the kernel replacement process. The main idea is to replace Gaussian kernels with minimal modifications, reusing the existing parameters of 3D Gaussian to preserve the original framework’s mathematical structure. This allows us to maintain the accumulation and blending operations without alteration. We begin by reformulating the Gaussian kernel in terms of the Mahalanobis distance \(D_M\):
\begin{gather}
    G(\mathbf{x}) = \exp \left( -\frac{1}{2} D_M^2 \right), \\
    D_M = \sqrt{(\mathbf{x} - \boldsymbol{\mu})^\top \Sigma^{-1} (\mathbf{x} - \boldsymbol{\mu})}.
\end{gather}

Here, \(D_M\) quantifies the distance between a point \(\mathbf{x}\) and the center \(\boldsymbol{\mu}\), scaled by the covariance matrix \(\Sigma\). Although the exponential decay in Gaussian kernels facilitates smooth blending across surfaces, it also suppresses high-frequency information, resulting in blurred edges and loss of fine details.

Recognizing that the kernel’s behavior is governed by its attenuation function, we can alter this behavior by applying a different attenuation model to \(D_M\). To address the limitations of Gaussian kernels, we replace the exponential decay with a linear attenuation function:
\begin{equation}
    L(\mathbf{x}) = \max\left(0, 1 - D_M \right).
\end{equation}

The linear kernel attenuates proportionally to the Mahalanobis distance \(D_M\), concentrating each splat’s influence within a confined region. This controlled attenuation preserves high-frequency details more effectively, reduces blending artifacts, and produces clearer reconstructions compared to the smoother Gaussian kernel.

The 2D projection of the linear kernel onto the image plane is given by:
\begin{gather}
    L'(\mathbf{x}') = \max\left(0, 1 - D'_M \right), \\
    D'_M = \sqrt{(\mathbf{x}' - \boldsymbol{\mu}')^\top \Sigma_{2D}^{-1} (\mathbf{x}' - \boldsymbol{\mu}')}.
\end{gather}

By preserving the underlying mathematical structure of the original 3DGS framework, the linear kernel integrates seamlessly into the existing pipeline. This substitution requires no changes to the accumulation and blending operations and significantly enhances rendering in high-frequency regions. As a result, 3DLS achieves sharper transitions, improved visual fidelity in intricate scenes, and faster rendering speeds.

\subsection{Training Optimization}

While the linear kernel enhances high-frequency detail capture, its distinct attenuation behavior introduces challenges during training. Specifically, its reduced coverage compared to Gaussian kernels can affect the representation of larger structures. Furthermore, the uniform gradient magnitude of the linear kernel may cause instability and slow convergence. To overcome these issues, we propose two optimization techniques: \textit{Distribution Alignment (DA)} and \textit{Adaptive Gradient Scaling (AGS)}, which ensure stable and efficient integration of the linear kernel into the 3DGS framework.

\begin{table*}[t]
\centering
\resizebox{\textwidth}{!}{\usebox{\tableQuantitative}}
\caption{Quantitative comparison demonstrating the superior performance of 3DLS on the Mip-NeRF360~\cite{Barron2022} and Tanks\&Temples datasets~\cite{Knapitsch2017}, with competitive results on Deep Blending~\cite{Hedman2018}. ``AA'' denotes the anti-aliasing method proposed by Mip-Splatting~\cite{Yu2024}. Results sourced from previous papers are marked with \textdagger.}
\vspace{-1em}
\label{tab:benchmark}
\end{table*}

\subsubsection{Distribution Alignment (DA)}
\label{section:distribution_alignment}

Replacing the Gaussian kernel with a linear kernel results in a narrower spread, which can limit its ability to accurately represent larger structures and potentially omit small but significant details. To ensure seamless integration within the 3DGS framework, we introduce an alignment factor \( \lambda \) to scale the Mahalanobis distance:
\begin{equation}
    L(\mathbf{x}) = \max\left(0, 1 - \frac{D_M}{\lambda} \right).
\end{equation}

The alignment factor \( \lambda \) adjusts the effective spread of the linear kernel to approximate that of the Gaussian kernel, ensuring that the total coverage area remains consistent. This alignment prevents the loss of detail in high-frequency regions, enabling the linear kernel to accurately render sharp edges and intricate textures without excessively reducing its footprint.

\begin{figure}[t]
\centering

\begin{minipage}[c]{0.95\linewidth}
\begin{minipage}[c]{0.05\linewidth}
\centering
(a)
\end{minipage}
\begin{minipage}[c]{0.9\linewidth}
\centering
\includegraphics[width=\linewidth]{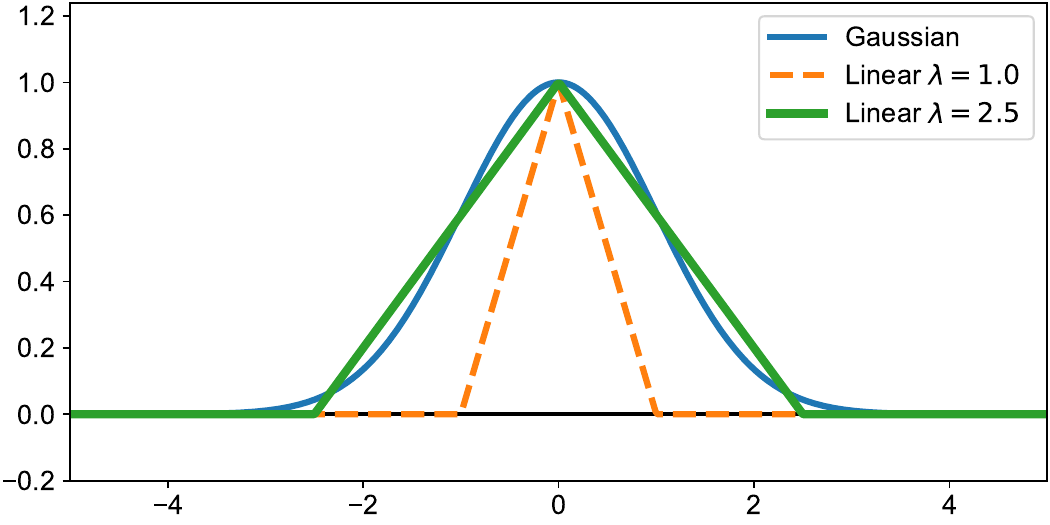}
\end{minipage}
\end{minipage}

\vspace*{0.2cm}

\begin{minipage}[c]{0.95\linewidth}
\begin{minipage}[c]{0.05\linewidth}
\centering
(b)
\end{minipage}
\begin{minipage}[c]{0.9\linewidth}
\centering
\includegraphics[width=\linewidth]{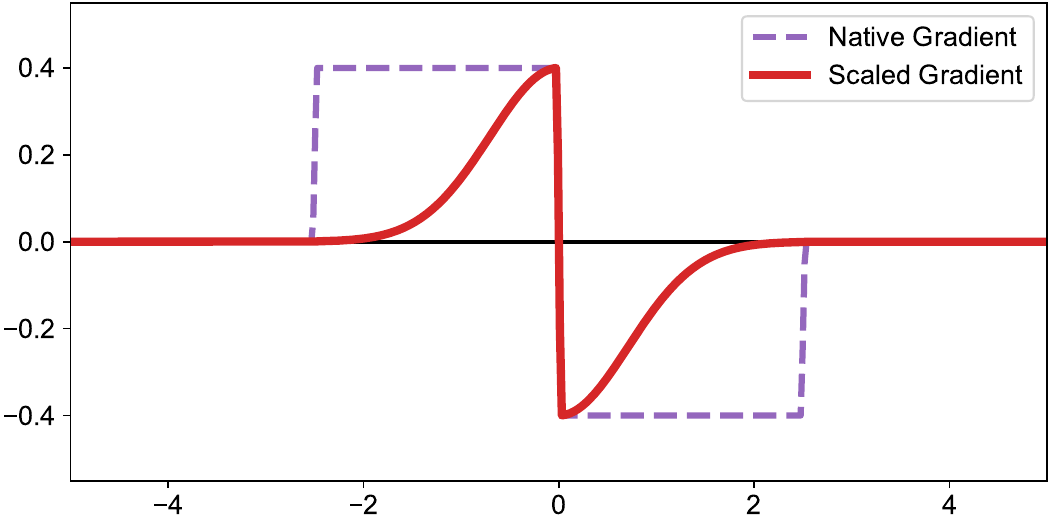}
\end{minipage}
\end{minipage}

\vspace{-0.5em}
\caption{(a) \textit{Distribution Alignment} (DA) adjusts the linear kernel to align with the coverage of the Gaussian kernel. (b) \textit{Adaptive Gradient Scaling} (AGS) smooths gradients, enhancing training stability and convergence.}
\vspace{-1em}
\label{fig:distribution_alignment}
\label{fig:gradient_scaling}
\end{figure}

We visualize the effect of \( \lambda \) on the spread of the linear kernel in Figure~\ref{fig:distribution_alignment}. The native spread concentrates the kernel’s influence more tightly, whereas applying \( \lambda = 2.5 \) broadens the spread to match the Gaussian kernel’s coverage, ensuring consistent detail capture across regions.

Incorporating \( \lambda \) directly into the Mahalanobis distance maintains consistent attenuation behavior between both kernels. This ensures that, despite their different mathematical forms, the transition between Gaussian and linear kernels remains smooth. Consequently, the linear kernel achieves enhanced detail capture while preserving the computational efficiency and compatibility of the 3DGS framework.

\subsubsection{Adaptive Gradient Scaling (AGS)}
\label{section:adaptive_gradient_scaling}

The linear kernel enhances the rendering of high-frequency details by providing sharper attenuation than the Gaussian kernel. However, the uniform gradient magnitude within the linear kernel’s support can introduce instability during training. Specifically, these uniform gradients may lead to disproportionate parameter updates from distant points, hindering convergence.

To address this, we propose an \textit{Adaptive Gradient Scaling (AGS)} strategy that modulates gradient magnitudes based on the Mahalanobis distance. This ensures that points farther from the kernel center have reduced influence, leading to more stable parameter updates. The scaling function is defined as:
\begin{equation}
    \omega(D'_M) = \exp\left( -{D'_M}^2 \right),
\end{equation}
where \( D'_M \) represents the Mahalanobis distance from the kernel center in 2D space.

\textit{AGS} improves training dynamics by adjusting gradient magnitudes proportionally to the point's distance. The gradient update for a parameter \( \theta \) becomes:
\begin{equation}
    \Delta \theta \propto -\nabla_\theta L'(\mathbf{x}') \cdot \omega(D'_M),
\end{equation}
where \( \omega(D'_M) \) serves as a scaling factor, prioritizing updates for regions with the greatest impact on the rendered image.

Figure~\ref{fig:gradient_scaling} illustrates the native and scaled gradient curves of the linear kernel. As shown, the native gradient remains constant across the kernel’s support, contrary to the expected behavior where distant points should exert less influence. Our \textit{AGS} resolves this issue, ensuring that the farther a pixel is from the center of the splat, the less impact it has on optimization.

Empirically, \textit{AGS} enhances both training stability and rendering performance. The model converges more reliably and captures fine details more effectively. Additionally, it mitigates issues such as overshooting and oscillations, which can occur with uniformly large gradients.

In summary, \textit{AGS} addresses the gradient-related challenges of the linear kernel, resulting in more stable optimization and improved rendering quality. This demonstrates how careful modifications to the optimization process can enhance model performance without compromising theoretical soundness or computational efficiency.

\begin{figure*}[t]
\centering
\includegraphics[width=\linewidth]{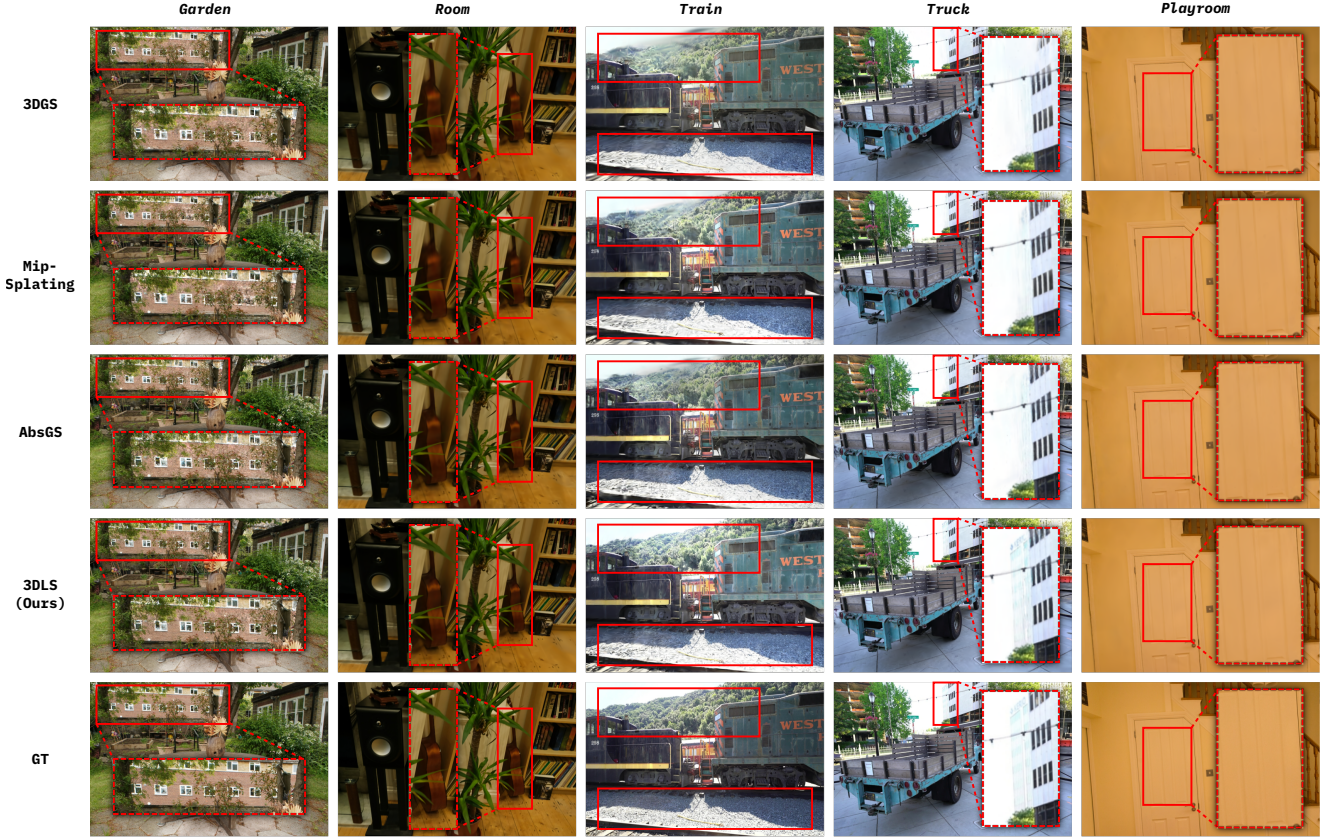}
\vspace{-1.5em}
\caption{Qualitative results demonstrate that our method excels in capturing high-frequency details, fine structures, and sharp transitions, resulting in higher-fidelity reconstructions.}
\vspace{-1em}
\label{fig:qualitative_results}
\end{figure*}


\section{Experiments}
\label{section:experiments}

In this section, we rigorously evaluate the performance and effectiveness of our proposed 3DLS approach, particularly emphasizing its advantages over Gaussian-based splatting methods in capturing high-frequency details. We conduct comprehensive benchmarks against existing SOTA methods across diverse datasets, incorporating both quantitative and qualitative assessments to highlight improvements in visual fidelity, efficiency, and computational performance.

\subsection{Implementation}

To evaluate the performance of our linear kernel, we benchmark it against several SOTA methods, including 3DGS~\cite{Kerbl2023}, 2DGS~\cite{Huang2024}, Mip-Splatting~\cite{Yu2024}, and AbsGS~\cite{Ye2024a}, using the Mip-NeRF360~\cite{Barron2022}, Tanks\&Temples~\cite{Knapitsch2017}, and Deep Blending~\cite{Hedman2018} datasets. For additional comparison, we include an anti-aliased version of our model (3DLS+AA) using the technique from Mip-Splatting. Our implementation is built on the \texttt{gsplat}~\cite{Ye2024} v1.3 codebase, which provides reference implementations for these baseline methods. To support the linear kernel and \textit{AGS} within this framework, we developed custom CUDA rasterization kernels optimized for both forward rendering and backpropagation.

To balance pixel-level accuracy with visual fidelity in textures and high-frequency details, we use a loss function that combines L1, L2, and SSIM~\cite{Wang2004} losses with weights of 6:2:2. Additionally, we apply a densification strategy based on 3DGS but with empirically adjusted thresholds tailored to optimize linear kernel performance. Specifically, we set the 3D growth threshold to 0.006 (from 0.01), the 3D prune threshold to 0.4 (from 0.1), and the opacity prune threshold to 0.025 (from 0.005). All other hyperparameters and settings are kept consistent with the baseline 3DGS to ensure a fair comparison.

\subsection{Evaluation}

We evaluate the performance of our proposed 3DLS method across four key aspects: quantitative comparisons with baseline methods, qualitative assessments to demonstrate visual improvements, an ablation study to analyze the contributions of each component, and an efficiency analysis to assess performance differences.

\begin{table*}[t]
\centering
\resizebox{\linewidth}{!}{\usebox{\tableAblation}}
\caption{Ablation study showing the impact of each component in our proposed method. Combining all components yields improved quantitative performance while keeping the number of primitives (N) at a manageable level.}
\vspace{-1em}
\label{tab:ablation}
\end{table*}

\begin{table}[t]
\centering
\resizebox{\linewidth}{!}{\usebox{\tablePerformance}}
\caption{3DLS achieves significantly higher rendering FPS with minimal impact on memory usage compared to other methods. ``Mip-Sp'' denotes Mip-Splatting.}
\vspace{-1em}
\label{tab:performance}
\end{table}

\vspace{-0.5em}
\subsubsection{Results Comparisons}

Table \ref{tab:benchmark} provides a comprehensive comparison between our method and existing radiance field rendering techniques, evaluated across key metrics: SSIM (Structural Similarity)~\cite{Wang2004}, PSNR (Peak Signal-to-Noise Ratio), and LPIPS (Learned Perceptual Image Patch Similarity)~\cite{Zhang2018} using VGG as the backbone. The results show consistent improvements across datasets, underscoring the robustness and versatility of our linear kernel approach.

\noindent
\textbf{Mip-NeRF360.} On the Mip-NeRF360 dataset~\cite{Barron2022}, our method achieves top-tier performance, outperforming 3DGS-based methods in both SSIM and PSNR. Additionally, we achieve the second-best LPIPS score, closely trailing AbsGS, which excels in perceptual quality. These results highlight our method’s ability to balance structural accuracy and visual fidelity, effectively capturing intricate transitions and preserving sharp details in high-frequency scenes.

\noindent
\textbf{Tanks\&Temples.} For the Tanks\&Temples~\cite{Knapitsch2017} dataset, our method outperforms all competing approaches across all three metrics. This dataset presents particular challenges due to its complex outdoor scenes and extensive view variations, and our results emphasize the linear kernel’s capability to handle these complexities without compromising visual clarity.

\noindent
\textbf{Deep Blending.} On the Deep Blending~\cite{Hedman2018} dataset, characterized by smoother, more continuous surfaces, our method achieves competitive SSIM and PSNR scores and secures the best LPIPS result. Although the advantage of the linear kernel is less pronounced in scenes with fewer high-frequency regions, our approach continues to excel in capturing perceptual details and minimizing visual artifacts.

In summary, our linear kernel consistently outperforms baselines across diverse datasets and metrics. The addition of anti-aliasing further enhances our results, providing stability and robustness across varied conditions. These findings validate the effectiveness of our approach in achieving a balanced trade-off between sharpness, fidelity, and perceptual quality across a range of scenarios.

\subsubsection{Qualitative Comparisons}

Figure \ref{fig:qualitative_results} presents a qualitative comparison between our method and current SOTA approaches. In the Garden scene, our method produces significantly fewer artifacts on the textured background wall. For sharp transitions, such as the window frames, our approach maintains a clear separation between the windows and the wall. In the Room scene, the black edge of the guitar is fully restored without blurring, underscoring our method’s ability to effectively preserve sharp transitions.

The advantages of our approach become even more evident in the Train scene. For the foreground railway ballast (gravel), both our method and AbsGS successfully reconstruct high-frequency details with minimal blurring. However, in the background mountain forests, our method is the only one to achieve reconstruction without noticeable blur. In the Truck scene, we accurately capture the fine details of the background building, where other methods struggle to reach this level of precision. Finally, in the Playroom scene, our method excels at preserving straight edges, such as the grooves on the door, where other methods introduce varying degrees of blurring.

Overall, our approach consistently surpasses existing methods by reducing artifacts, maintaining sharp transitions, and achieving higher fidelity in both foreground and background details across diverse scenes.

\begin{figure*}[t]
\centering
\includegraphics[width=0.95\linewidth]{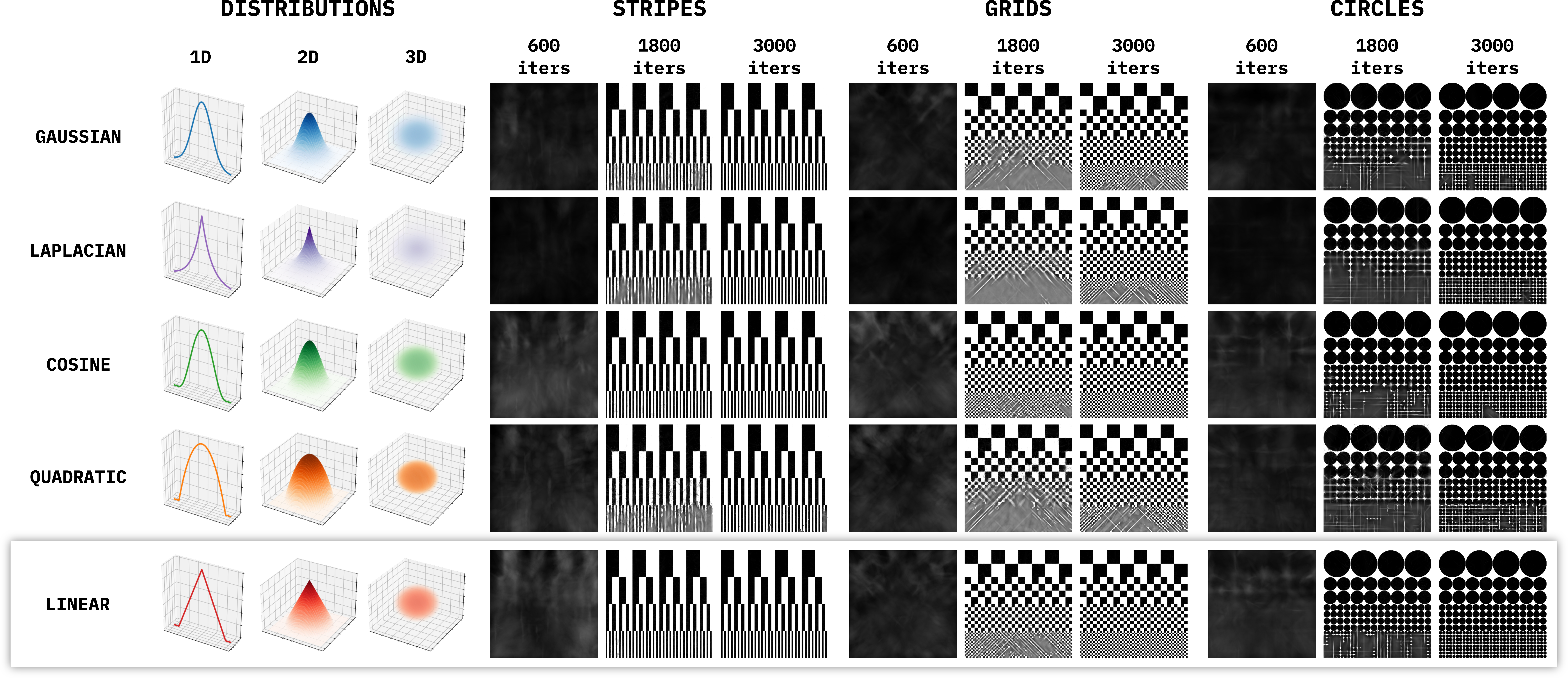}
\caption{Evaluation of different kernels on complex patterns to simulate challenging cases. Results indicate that the linear kernel excels in reconstructing high-frequency regions.}
\vspace{-1em}
\label{fig:distri_patterns}
\end{figure*}

\subsubsection{Ablation Study}

Table \ref{tab:ablation} illustrates the impact of our proposed components: the linear kernel (\textit{LK}), \textit{Distribution Alignment (DA)}, and \textit{Adaptive Gradient Scaling (AGS)} on splatting-based rendering. We evaluated our method on two representative scenes, Kitchen (indoor) and Train (outdoor), to capture a range of real-world complexities. The baseline 3DGS, shown in the first row, serves as a reference for comparison.

\noindent
\textbf{Efficiency Gains with Linear Kernel:} Incorporating the linear kernel alone achieves performance comparable to the baseline 3DGS while significantly reducing the number of primitives (\(N\)), creating a more compact and efficient representation. This efficiency gain demonstrates the linear kernel’s ability to sustain high-quality reconstruction with fewer resources, advantageous for performance-critical applications.

\noindent
\textbf{Fidelity Boost from \textit{DA}:} Applying \textit{DA} further improves our method, surpassing the baseline 3DGS across all metrics. This enhancement, however, raises the primitive count (\(N\)), as \textit{DA} increases the spread of splats for broader coverage and finer detail capture. The improved SSIM and LPIPS scores indicate that \textit{DA} is instrumental in boosting visual fidelity, especially in complex scenes.

\noindent
\textbf{Efficiency-Fidelity Balance with \textit{AGS}:} Adding \textit{AGS} moderates the primitive count, bringing it closer to the baseline while retaining high SSIM and LPIPS scores and achieving a notable PSNR boost. By optimizing the primitive distribution in high-frequency areas, \textit{AGS} balances detail preservation with computational efficiency, avoiding excessive splat counts.

\vspace{-1em}
\subsubsection{Efficiency Analysis}

To assess practical performance differences, we profile the forward and backward FPS, memory usage, and number of primitives (\(N\)) of our proposed 3DLS method alongside established 3DGS methods, including the baseline 3DGS~\cite{Kerbl2023}, Mip-Splatting~\cite{Yu2024}, and AbsGS~\cite{Ye2024a}. Testing was conducted on three benchmark datasets across all views, with each scene undergoing five warm-up runs to ensure consistency. As shown in Table~\ref{tab:performance}, 3DLS achieves a substantial increase in rendering speed, with both forward and backward passes improving by over 30\% due to the computational efficiency of the linear kernel. Notably, while our approach results in higher primitive counts, similar to detail-intensive methods like AbsGS, memory usage remains low. This efficiency is attributed to the separation of linear splats, which, unlike Gaussian splats, reduce the need for intensive per-pixel blending and thus minimize memory demands. Overall, 3DLS offers significant gains in rendering speed with minimal memory overhead, affirming the linear kernel's effectiveness and suitability for high-performance applications.

\section{Discussion}

\subsection{Exploring Kernel Distributions}
\label{section:kernel_distributions}

Our work introduces a new perspective on kernel functions in splatting-based rendering, demonstrating that kernel choice significantly affects reconstruction quality, particularly in high-frequency regions. We evaluated various kernels, including Gaussian, Laplacian, Cosine, Quadratic, and Linear, by fitting them to complex patterns such as stripes, checkerboards, and circles with varying frequencies (Figure~\ref{fig:distri_patterns}). This analysis revealed distinct behaviors for each kernel, with the linear kernel emerging as particularly effective for preserving high-frequency details.

Gaussian and Laplacian kernels, while effective for low-frequency patterns, face limitations in high-frequency regions due to their long tails, which cause blurring and overlapping artifacts. In contrast, shorter-tailed kernels, such as the raised cosine, are better suited for handling high-frequency textures and reducing artifacts in intricate regions. Conversely, the quadratic kernel’s parabolic falloff improves splat uniformity but requires splats to be placed closer together, hindering spatial clarity in high-frequency scenes.

The linear kernel’s bounded support and gradual decay make it particularly effective across diverse patterns, providing high fidelity in high-frequency regions without noticeable artifacts. Its computational efficiency and ability to handle intricate details position it as a compelling choice for high-fidelity 3D splatting. While our work focuses on the linear kernel for its clear advantages, future research could explore hybrid models that dynamically adapt kernel choices based on scene complexity, potentially advancing splatting-based rendering further.

\subsection{Limitations}

While our linear kernel approach offers substantial performance gains, it has certain limitations. The densification threshold controlling primitive growth and pruning is set empirically; a systematic optimization across diverse scenes could enhance our results. Additionally, as our method is built on a 3DGS foundation, it may not fully exploit the linear kernel’s potential; architectural adaptations tailored to the linear kernel could yield further performance gains. Our approach excels in datasets with high-frequency details, where sharpness is critical, but is less impactful in smoother datasets with continuous surfaces. Future work on adaptive or hybrid kernels may enable consistently high-quality results across a broader range of scenes.


\section{Conclusion}
\label{section:conclusion}

This work introduced 3DLS, a novel approach that advances 3D reconstruction fidelity by addressing core limitations in traditional 3DGS methods. By utilizing linear kernels, 3DLS captures high-frequency details with exceptional accuracy, delivering superior performance across diverse datasets. Extensive experiments demonstrate that 3DLS consistently outperforms existing methods, particularly in scenes with intricate textures and fine details. Additionally, 3DLS achieves substantial gains in rendering speed with minimal memory overhead, making it highly suited to performance-critical applications. These findings underscore the importance of kernel design in splatting-based rendering, paving the way for further exploration of adaptive and hybrid kernels to enhance both fidelity and efficiency in 3D rendering systems.

{
    \small
    \bibliographystyle{ieeenat_fullname}
    \bibliography{main}
}

\clearpage
\setcounter{page}{1}
\maketitlesupplementary
\appendix

This supplementary document provides additional \textbf{technical and methodological details} of our work, complete \textbf{per-scene experimental results}, and \textbf{additional qualitative examples}. These materials aim to offer a more comprehensive demonstration of the proposed approach.

\vspace{-0.2em}
\section{Implementation Details}
\vspace{-0.2em}

This section outlines the hardware specifications for evaluations, the formulations of the loss functions, and the densification thresholds used in all experiments. These details are provided to ensure reproducibility and to give insights into the computational requirements of our approach.

\noindent
\textbf{Hardware Specifications.} 
All evaluations were conducted on a workstation equipped with an AMD Ryzen 9 5900X CPU, 32GB of RAM, and an NVIDIA GeForce RTX 3090 GPU. The software environment comprised PyTorch 2.1.0 with CUDA 11.8, running on Ubuntu 22.04 LTS.

\noindent
\textbf{Loss Functions.} 
We use a weighted combination of L1, L2, and SSIM losses to balance pixel-level accuracy and structural similarity, enhancing visual fidelity for diverse textures and high-frequency details. The total loss is defined as:
\begin{equation}
\mathcal{L}_{\text{total}} = \alpha \mathcal{L}_{\text{L1}} + \beta \mathcal{L}_{\text{L2}} + \gamma \mathcal{L}_{\text{SSIM}},
\end{equation}
where \(\alpha = 0.6\), \(\beta = 0.2\), and \(\gamma = 0.2\), reflecting a 6:2:2 ratio. The individual loss terms are defined as follows:
\begin{align}
\mathcal{L}_{\text{L1}} &= \frac{1}{N} \sum_{i=1}^{N} \lvert \hat{y}_i - y_i \rvert, \\
\mathcal{L}_{\text{L2}} &= \frac{1}{N} \sum_{i=1}^{N} (\hat{y}_i - y_i)^2, \\
\mathcal{L}_{\text{SSIM}} &= 1 - \text{SSIM}(\hat{y}, y).
\end{align}

This combined loss optimizes reconstruction quality by balancing precise pixel alignment, robust error handling, and perceptual fidelity across a variety of scenes.

\noindent
\textbf{Densification Thresholds.} 
The thresholds used for 3DGS-based methods and our proposed 3DLS methods are summarized in Table~\ref{tab:densification_thresholds}, covering gradient threshold ($\tau_{\nabla g}$), 2D/3D grow thresholds ($\tau_{s2g}$, $\tau_{s3g}$), 2D/3D prune thresholds ($\tau_{s2p}$, $\tau_{s3p}$), and opacity prune threshold ($\tau_{op}$).

\begin{table}[!hb]
\centering
\setlength{\heavyrulewidth}{0.06em}
\setlength{\lightrulewidth}{0.01em}
\setlength{\cmidrulewidth}{0.01em}
\setlength{\aboverulesep}{0.2em}
\setlength{\belowrulesep}{0.2em}
\renewcommand{\arraystretch}{0.5}
\resizebox{\linewidth}{!}{
\scriptsize
\begin{NiceTabular}{c|cccccc}
\toprule
\multirow{2}{*}{Method} & \multicolumn{6}{c}{Densification Parameters} \\
& $\tau_{\nabla g}$ & $\tau_{s2g}$ & $\tau_{s3g}$ & $\tau_{s2p}$ & $\tau_{s3p}$ & $\tau_{op}$ \\
\midrule
3DGS & 0.0002 & 0.05 & 0.01 & 0.15 & 0.1 & 0.005 \\
\midrule
3DLS (Ours) & 0.0002 & 0.05 & 0.006 & 0.15 & 0.4 & 0.025 \\
\bottomrule
\end{NiceTabular}}
\caption{Densification thresholds for 3DGS and 3DLS.}
\label{tab:densification_thresholds}
\end{table}

\section{Generalizing the 3DGS Kernel Function}

In Section~\ref{section:linear_kernel}, we describe replacing the Gaussian kernel with a Linear kernel. Here, we generalize the 3DGS kernel function to support other types of kernels.

We define an ellipsoid function \( E(\mathbf{x}) \), parameterized by the covariance matrix \(\Sigma\), as:
\begin{gather}
E(\mathbf{x}) = f\left( D(\mathbf{x}) \right), \\
D(\mathbf{x}) = \sqrt{\mathbf{x}^\top \Sigma^{-1} \mathbf{x}},
\end{gather}
where \( D(\mathbf{x}) \) is the Mahalanobis distance. By choosing different forms of the attenuation function \( f(D) \), various kernel types can be modeled:
\begin{align}
\text{Gaussian:} & \quad f(D) = \exp\left( -\tfrac{1}{2} D^2 \right), \\
\text{Laplacian:} & \quad f(D) = \exp\left( -D \right), \\
\text{Raised Cosine:} & \quad 
\begin{cases}
\frac{1}{2} \left[ 1 + \cos\left( \pi D \right) \right], & \text{if } D \leq 1, \\
0, & \text{if } D > 1,
\end{cases} \\
\text{Quadratic:} & \quad f(D) = \max(0, 1 - D^2), \\
\text{Linear:} & \quad f(D) = \max(0, 1 - D).
\end{align}

This formulation provides a unified framework for accommodating diverse kernel types, enhancing the flexibility and adaptability of the 3DGS framework.

Additionally, our proposed \textit{Distribution Alignment (DA)} method can be seamlessly applied to all the listed functions. This is achieved by normalizing \( D \) using a scaling factor \(\lambda\), defined as:
\begin{equation}
D_{\text{aligned}} = \frac{D}{\lambda}.
\end{equation}

Empirical tests show that \(\lambda = 2.5\) aligns Cosine and Linear kernels to the Gaussian distribution, \(\lambda = 6\) works for Quadratic, and \(\lambda = 1\) suffices for Laplacian. This scaling ensures consistent attenuation behavior across kernels, enhancing robustness for diverse scenarios.

\section{Per-Scene Experiment Results}

We present detailed per-scene experimental results across three datasets: Mip-NeRF360~\cite{Barron2022}, Tanks \& Temples~\cite{Knapitsch2017}, and Deep Blending~\cite{Hedman2018}. Our proposed 3DLS method, including its anti-aliased variant (3DLS+AA), is compared against state-of-the-art approaches such as 3DGS~\cite{Kerbl2023}, 2DGS~\cite{Huang2024}, Mip-Splatting~\cite{Yu2024}, and AbsGS~\cite{Ye2024a}. Table~\ref{tab:per_scene} reports results for structural similarity (SSIM), peak signal-to-noise ratio (PSNR), and learned perceptual image patch similarity (LPIPS), highlighting the per-scene performance of each method.

\begin{table*}
\centering

\caption*{\textbf{SSIM}}
\vspace{-0.5em}
\resizebox{\linewidth}{!}{
\begin{NiceTabular}{c|l|ccccccccc|c|cc|c|cc|c}[colortbl-like]
\toprule
\Block{2-1}{Step} & \Block{2-1}{Method} & \multicolumn{9}{c}{Mip-NeRF360} & \Block{2-1}{MEAN} &\multicolumn{2}{c}{Tanks\&Temples} & \Block{2-1}{MEAN} &\multicolumn{2}{c}{Deep Blending} & \Block{2-1}{MEAN} \\ 
 & & bicycle & garden & stump & room & counter & kitchen & bonsai & flowers & treehill &  & truck & train &  & drjohnson & playroom &  \\
\midrule
\Block{6-1}{7k} & 3DGS & 0.659 & \cellcolor{t_yellow} 0.830 & 0.720 & 0.892 & \cellcolor{t_yellow} 0.877 & \cellcolor{t_yellow} 0.902 & \cellcolor{t_yellow} 0.922 & 0.519 & 0.588 & 0.768 & \cellcolor{t_yellow} 0.848 & \cellcolor{t_orange} 0.711 & 0.780 & \cellcolor{t_orange} 0.865 & 0.895 & \cellcolor{t_yellow} 0.880 \\
 & 2DGS & 0.634 & 0.813 & 0.707 & \cellcolor{t_yellow} 0.893 & 0.872 & 0.895 & \cellcolor{t_yellow} 0.922 & 0.510 & 0.572 & 0.758 & 0.847 & 0.697 & 0.772 & \cellcolor{t_orange} 0.865 & \cellcolor{t_yellow} 0.896 & \cellcolor{t_yellow} 0.880 \\
 & Mip-Splatting & 0.656 & 0.823 & 0.717 & \cellcolor{t_yellow} 0.893 & \cellcolor{t_yellow} 0.877 & \cellcolor{t_yellow} 0.902 & \cellcolor{t_yellow} 0.922 & 0.516 & 0.586 & 0.766 & 0.845 & \cellcolor{t_yellow} 0.708 & 0.776 & \cellcolor{t_red} 0.867 & \cellcolor{t_yellow} 0.896 & \cellcolor{t_orange} 0.881 \\
 & AbsGS & \cellcolor{t_red} 0.713 & \cellcolor{t_red} 0.846 & \cellcolor{t_red} 0.763 & \cellcolor{t_yellow} 0.893 & \cellcolor{t_orange} 0.884 & \cellcolor{t_orange} 0.905 & \cellcolor{t_yellow} 0.922 & \cellcolor{t_red} 0.571 & \cellcolor{t_red} 0.608 & \cellcolor{t_red} 0.789 & \cellcolor{t_orange} 0.859 & \cellcolor{t_orange} 0.711 & \cellcolor{t_yellow} 0.785 & 0.848 & 0.895 & 0.872 \\
 & 3DLS (Ours) & \cellcolor{t_yellow} 0.693 & \cellcolor{t_orange} 0.841 & \cellcolor{t_yellow} 0.746 & \cellcolor{t_orange} 0.900 & \cellcolor{t_orange} 0.884 & \cellcolor{t_red} 0.908 & \cellcolor{t_orange} 0.928 & \cellcolor{t_yellow} 0.551 & \cellcolor{t_yellow} 0.602 & \cellcolor{t_yellow} 0.784 & \cellcolor{t_orange} 0.859 & \cellcolor{t_red} 0.728 & \cellcolor{t_orange} 0.793 & 0.863 & \cellcolor{t_orange} 0.898 & \cellcolor{t_orange} 0.881 \\
 & 3DLS+AA (Ours) & \cellcolor{t_orange} 0.696 & \cellcolor{t_orange} 0.841 & \cellcolor{t_orange} 0.748 & \cellcolor{t_red} 0.901 & \cellcolor{t_red} 0.885 & \cellcolor{t_red} 0.908 & \cellcolor{t_red} 0.929 & \cellcolor{t_orange} 0.554 & \cellcolor{t_orange} 0.606 & \cellcolor{t_orange} 0.785 & \cellcolor{t_red} 0.862 & \cellcolor{t_red} 0.728 & \cellcolor{t_red} 0.795 & \cellcolor{t_yellow} 0.864 & \cellcolor{t_red} 0.899 & \cellcolor{t_red} 0.882 \\
\midrule
\Block{6-1}{30k} & 3DGS & 0.760 & 0.867 & 0.773 & 0.920 & 0.908 & \cellcolor{t_yellow} 0.928 & 0.942 & 0.603 & 0.634 & \cellcolor{t_yellow} 0.815 & 0.881 & 0.814 & 0.847 & \cellcolor{t_orange} 0.902 & \cellcolor{t_orange} 0.908 & \cellcolor{t_orange} 0.905 \\
 & 2DGS & 0.744 & 0.851 & 0.762 & 0.917 & 0.902 & 0.923 & 0.939 & 0.592 & 0.618 & 0.805 & 0.876 & 0.784 & 0.830 & \cellcolor{t_yellow} 0.901 & \cellcolor{t_orange} 0.908 & \cellcolor{t_yellow} 0.904 \\
 & Mip-Splatting & 0.762 & 0.864 & 0.770 & 0.920 & 0.908 & \cellcolor{t_yellow} 0.928 & 0.942 & 0.600 & \cellcolor{t_orange} 0.637 & \cellcolor{t_yellow} 0.815 & 0.882 & 0.813 & 0.848 & \cellcolor{t_red} 0.903 & \cellcolor{t_red} 0.909 & \cellcolor{t_red} 0.906 \\
 & AbsGS & \cellcolor{t_yellow} 0.772 & \cellcolor{t_red} 0.873 & \cellcolor{t_red} 0.783 & \cellcolor{t_yellow} 0.923 & \cellcolor{t_red} 0.914 & \cellcolor{t_orange} 0.929 & \cellcolor{t_yellow} 0.944 & \cellcolor{t_red} 0.636 & 0.625 & \cellcolor{t_orange} 0.822 & \cellcolor{t_orange} 0.887 & \cellcolor{t_yellow} 0.821 & \cellcolor{t_yellow} 0.854 & 0.890 & 0.906 & 0.898 \\
 & 3DLS (Ours) & \cellcolor{t_orange} 0.773 & \cellcolor{t_yellow} 0.869 & \cellcolor{t_yellow} 0.778 & \cellcolor{t_orange} 0.924 & \cellcolor{t_yellow} 0.911 & \cellcolor{t_red} 0.931 & \cellcolor{t_orange} 0.945 & \cellcolor{t_yellow} 0.627 & \cellcolor{t_yellow} 0.635 & \cellcolor{t_orange} 0.822 & \cellcolor{t_yellow} 0.885 & \cellcolor{t_orange} 0.825 & \cellcolor{t_orange} 0.855 & 0.897 & \cellcolor{t_yellow} 0.907 & 0.902 \\
 & 3DLS+AA (Ours) & \cellcolor{t_red} 0.776 & \cellcolor{t_orange} 0.872 & \cellcolor{t_orange} 0.779 & \cellcolor{t_red} 0.925 & \cellcolor{t_orange} 0.913 & 0.924 & \cellcolor{t_red} 0.946 & \cellcolor{t_orange} 0.631 & \cellcolor{t_red} 0.640 & \cellcolor{t_red} 0.823 & \cellcolor{t_red} 0.890 & \cellcolor{t_red} 0.830 & \cellcolor{t_red} 0.860 & 0.899 & \cellcolor{t_orange} 0.908 & \cellcolor{t_yellow} 0.904 \\
\bottomrule
\end{NiceTabular}}

\vspace{2em}

\caption*{\textbf{PSNR}}
\vspace{-0.5em}
\resizebox{\linewidth}{!}{
\begin{NiceTabular}{c|l|ccccccccc|c|cc|c|cc|c}[colortbl-like]
\toprule
\Block{2-1}{Step} & \Block{2-1}{Method} & \multicolumn{9}{c}{Mip-NeRF360} & \Block{2-1}{MEAN} &\multicolumn{2}{c}{Tanks\&Temples} & \Block{2-1}{MEAN} &\multicolumn{2}{c}{Deep Blending} & \Block{2-1}{MEAN} \\ 
 & & bicycle & garden & stump & room & counter & kitchen & bonsai & flowers & treehill &  & truck & train &  & drjohnson & playroom &  \\
\midrule
\Block{6-1}{7k} & 3DGS & 23.450 & 26.260 & 25.570 & 29.140 & 27.160 & 29.050 & 29.710 & 20.440 & \cellcolor{t_yellow} 22.120 & 25.880 & 23.910 & \cellcolor{t_yellow} 19.500 & \cellcolor{t_yellow} 21.700 & \cellcolor{t_yellow} 27.120 & \cellcolor{t_yellow} 29.300 & 28.210 \\
 & 2DGS & 23.010 & 25.970 & 25.290 & \cellcolor{t_yellow} 29.250 & 27.130 & 28.950 & 29.620 & 20.210 & 21.780 & 25.690 & 23.950 & 18.850 & 21.400 & \cellcolor{t_red} 27.460 & \cellcolor{t_red} 29.440 & \cellcolor{t_red} 28.450 \\
 & Mip-Splatting & 23.420 & 26.080 & 25.520 & 29.120 & 27.190 & \cellcolor{t_yellow} 29.090 & 29.750 & 20.420 & 22.070 & 25.850 & 23.930 & 19.430 & 21.680 & \cellcolor{t_orange} 27.210 & \cellcolor{t_yellow} 29.300 & \cellcolor{t_yellow} 28.260 \\
 & AbsGS & \cellcolor{t_red} 23.820 & \cellcolor{t_red} 26.620 & \cellcolor{t_red} 26.380 & 28.630 & \cellcolor{t_yellow} 27.260 & 28.920 & \cellcolor{t_yellow} 29.770 & \cellcolor{t_red} 21.050 & 22.020 & \cellcolor{t_yellow} 26.050 & \cellcolor{t_yellow} 24.240 & 19.150 & \cellcolor{t_yellow} 21.700 & 25.520 & 29.000 & 27.260 \\
 & 3DLS (Ours) & \cellcolor{t_yellow} 23.640 & \cellcolor{t_orange} 26.480 & \cellcolor{t_yellow} 25.990 & \cellcolor{t_orange} 29.420 & \cellcolor{t_orange} 27.380 & \cellcolor{t_orange} 29.610 & \cellcolor{t_orange} 29.950 & \cellcolor{t_yellow} 20.900 & \cellcolor{t_orange} 22.360 & \cellcolor{t_orange} 26.190 & \cellcolor{t_orange} 24.280 & \cellcolor{t_red} 19.790 & \cellcolor{t_orange} 22.030 & 27.020 & \cellcolor{t_orange} 29.410 & 28.220 \\
 & 3DLS+AA (Ours) & \cellcolor{t_orange} 23.670 & \cellcolor{t_yellow} 26.460 & \cellcolor{t_orange} 26.030 & \cellcolor{t_red} 29.510 & \cellcolor{t_red} 27.430 & \cellcolor{t_red} 29.620 & \cellcolor{t_red} 30.080 & \cellcolor{t_orange} 20.960 & \cellcolor{t_red} 22.470 & \cellcolor{t_red} 26.250 & \cellcolor{t_red} 24.430 & \cellcolor{t_orange} 19.750 & \cellcolor{t_red} 22.090 & 27.100 & \cellcolor{t_red} 29.440 & \cellcolor{t_orange} 28.270 \\
\midrule
\Block{6-1}{30k} & 3DGS & 25.110 & \cellcolor{t_yellow} 27.410 & 26.600 & \cellcolor{t_orange} 31.480 & 28.990 & \cellcolor{t_yellow} 31.400 & 32.170 & 21.560 & \cellcolor{t_yellow} 22.510 & 27.470 & 25.360 & \cellcolor{t_yellow} 21.880 & 23.620 & 28.990 & 29.990 & 29.490 \\
 & 2DGS & 24.880 & 26.990 & 26.310 & 31.230 & 28.770 & 31.220 & 31.920 & 21.370 & 22.150 & 27.200 & 25.220 & 20.560 & 22.890 & \cellcolor{t_red} 29.250 & \cellcolor{t_orange} 30.080 & \cellcolor{t_red} 29.670 \\
 & Mip-Splatting & \cellcolor{t_yellow} 25.170 & 27.300 & 26.510 & \cellcolor{t_red} 31.530 & \cellcolor{t_orange} 29.070 & \cellcolor{t_orange} 31.440 & \cellcolor{t_yellow} 32.320 & \cellcolor{t_yellow} 21.600 & \cellcolor{t_red} 22.620 & \cellcolor{t_yellow} 27.510 & \cellcolor{t_yellow} 25.520 & 21.780 & \cellcolor{t_yellow} 23.650 & \cellcolor{t_orange} 29.190 & \cellcolor{t_red} 30.140 & \cellcolor{t_orange} 29.660 \\
 & AbsGS & \cellcolor{t_orange} 25.190 & \cellcolor{t_red} 27.550 & \cellcolor{t_red} 26.740 & 31.200 & \cellcolor{t_orange} 29.070 & 31.240 & \cellcolor{t_orange} 32.340 & 21.500 & 21.850 & 27.410 & \cellcolor{t_orange} 25.530 & 21.330 & 23.430 & 27.800 & 29.730 & 28.770 \\
 & 3DLS (Ours) & \cellcolor{t_orange} 25.190 & 27.400 & \cellcolor{t_yellow} 26.610 & \cellcolor{t_yellow} 31.390 & \cellcolor{t_yellow} 29.060 & \cellcolor{t_red} 31.780 & \cellcolor{t_yellow} 32.320 & \cellcolor{t_orange} 21.900 & 22.450 & \cellcolor{t_orange} 27.570 & 25.490 & \cellcolor{t_orange} 21.930 & \cellcolor{t_orange} 23.710 & 28.870 & \cellcolor{t_yellow} 30.020 & 29.440 \\
 & 3DLS+AA (Ours) & \cellcolor{t_red} 25.270 & \cellcolor{t_orange} 27.500 & \cellcolor{t_orange} 26.630 & \cellcolor{t_red} 31.530 & \cellcolor{t_red} 29.250 & 31.300 & \cellcolor{t_red} 32.580 & \cellcolor{t_red} 22.040 & \cellcolor{t_orange} 22.570 & \cellcolor{t_red} 27.630 & \cellcolor{t_red} 25.790 & \cellcolor{t_red} 21.980 & \cellcolor{t_red} 23.890 & \cellcolor{t_yellow} 29.010 & 29.990 & \cellcolor{t_yellow} 29.500 \\
\bottomrule
\end{NiceTabular}}

\vspace{2em}

\caption*{\textbf{LPIPS}}
\vspace{-0.5em}
\resizebox{\linewidth}{!}{
\begin{NiceTabular}{c|l|ccccccccc|c|cc|c|cc|c}[colortbl-like]
\toprule
\Block{2-1}{Step} & \Block{2-1}{Method} & \multicolumn{9}{c}{Mip-NeRF360} & \Block{2-1}{MEAN} &\multicolumn{2}{c}{Tanks\&Temples} & \Block{2-1}{MEAN} &\multicolumn{2}{c}{Deep Blending} & \Block{2-1}{MEAN} \\ 
 & & bicycle & garden & stump & room & counter & kitchen & bonsai & flowers & treehill &  & truck & train &  & drjohnson & playroom &  \\
\midrule
\Block{6-1}{7k} & 3DGS & 0.335 & \cellcolor{t_yellow} 0.162 & 0.297 & 0.264 & \cellcolor{t_yellow} 0.249 & 0.164 & 0.238 & \cellcolor{t_yellow} 0.419 & \cellcolor{t_yellow} 0.416 & 0.283 & 0.203 & 0.332 & 0.268 & \cellcolor{t_orange} 0.328 & 0.295 & 0.312 \\
 & 2DGS & 0.355 & 0.184 & 0.311 & 0.267 & 0.255 & 0.175 & \cellcolor{t_yellow} 0.237 & 0.423 & 0.428 & 0.293 & 0.207 & 0.354 & 0.281 & 0.333 & 0.297 & 0.315 \\
 & Mip-Splatting & 0.339 & 0.173 & 0.301 & 0.265 & \cellcolor{t_yellow} 0.249 & \cellcolor{t_yellow} 0.163 & 0.239 & 0.424 & 0.418 & 0.286 & 0.211 & 0.335 & 0.273 & \cellcolor{t_red} 0.327 & 0.295 & \cellcolor{t_yellow} 0.311 \\
 & AbsGS & \cellcolor{t_red} 0.256 & \cellcolor{t_red} 0.135 & \cellcolor{t_red} 0.234 & \cellcolor{t_yellow} 0.249 & \cellcolor{t_orange} 0.233 & \cellcolor{t_orange} 0.153 & \cellcolor{t_orange} 0.221 & \cellcolor{t_red} 0.361 & \cellcolor{t_red} 0.361 & \cellcolor{t_red} 0.245 & \cellcolor{t_yellow} 0.183 & \cellcolor{t_yellow} 0.325 & \cellcolor{t_yellow} 0.254 & 0.348 & \cellcolor{t_yellow} 0.290 & 0.319 \\
 & 3DLS (Ours) & \cellcolor{t_yellow} 0.281 & \cellcolor{t_orange} 0.138 & \cellcolor{t_yellow} 0.253 & \cellcolor{t_orange} 0.245 & \cellcolor{t_orange} 0.233 & \cellcolor{t_red} 0.151 & \cellcolor{t_red} 0.219 & \cellcolor{t_orange} 0.376 & \cellcolor{t_orange} 0.369 & \cellcolor{t_yellow} 0.252 & \cellcolor{t_orange} 0.181 & \cellcolor{t_orange} 0.307 & \cellcolor{t_orange} 0.244 & \cellcolor{t_yellow} 0.330 & \cellcolor{t_orange} 0.285 & \cellcolor{t_orange} 0.307 \\
 & 3DLS+AA (Ours) & \cellcolor{t_orange} 0.278 & \cellcolor{t_orange} 0.138 & \cellcolor{t_orange} 0.251 & \cellcolor{t_red} 0.244 & \cellcolor{t_red} 0.232 & \cellcolor{t_red} 0.151 & \cellcolor{t_red} 0.219 & \cellcolor{t_orange} 0.376 & \cellcolor{t_orange} 0.369 & \cellcolor{t_orange} 0.251 & \cellcolor{t_red} 0.179 & \cellcolor{t_red} 0.306 & \cellcolor{t_red} 0.243 & \cellcolor{t_orange} 0.328 & \cellcolor{t_red} 0.282 & \cellcolor{t_red} 0.305 \\
\midrule
\Block{6-1}{30k} & 3DGS & \cellcolor{t_yellow} 0.208 & 0.107 & 0.214 & 0.221 & 0.201 & \cellcolor{t_yellow} 0.127 & 0.208 & \cellcolor{t_yellow} 0.336 & \cellcolor{t_yellow} 0.323 & \cellcolor{t_yellow} 0.216 & 0.148 & 0.207 & 0.178 & \cellcolor{t_orange} 0.246 & 0.248 & 0.247 \\
 & 2DGS & 0.235 & 0.130 & 0.232 & 0.228 & 0.212 & 0.137 & 0.211 & 0.349 & 0.347 & 0.231 & 0.161 & 0.244 & 0.203 & \cellcolor{t_yellow} 0.247 & 0.251 & 0.249 \\
 & Mip-Splatting & 0.216 & 0.115 & 0.219 & 0.221 & 0.202 & 0.128 & 0.207 & 0.343 & 0.329 & 0.220 & 0.152 & 0.210 & 0.181 & \cellcolor{t_red} 0.243 & \cellcolor{t_yellow} 0.246 & \cellcolor{t_yellow} 0.245 \\
 & AbsGS & \cellcolor{t_red} 0.167 & \cellcolor{t_red} 0.096 & \cellcolor{t_red} 0.188 & \cellcolor{t_red} 0.203 & \cellcolor{t_red} 0.185 & \cellcolor{t_red} 0.120 & \cellcolor{t_red} 0.190 & \cellcolor{t_red} 0.257 & \cellcolor{t_red} 0.268 & \cellcolor{t_red} 0.186 & \cellcolor{t_red} 0.128 & \cellcolor{t_yellow} 0.189 & \cellcolor{t_orange} 0.158 & 0.256 & \cellcolor{t_orange} 0.242 & 0.249 \\
 & 3DLS (Ours) & \cellcolor{t_orange} 0.179 & \cellcolor{t_yellow} 0.100 & \cellcolor{t_yellow} 0.194 & \cellcolor{t_yellow} 0.206 & \cellcolor{t_yellow} 0.190 & \cellcolor{t_red} 0.120 & \cellcolor{t_yellow} 0.195 & \cellcolor{t_orange} 0.299 & \cellcolor{t_orange} 0.284 & \cellcolor{t_orange} 0.196 & \cellcolor{t_yellow} 0.136 & \cellcolor{t_orange} 0.183 & \cellcolor{t_yellow} 0.160 & \cellcolor{t_yellow} 0.247 & \cellcolor{t_orange} 0.242 & \cellcolor{t_orange} 0.244 \\
 & 3DLS+AA (Ours) & \cellcolor{t_orange} 0.179 & \cellcolor{t_orange} 0.098 & \cellcolor{t_orange} 0.191 & \cellcolor{t_orange} 0.204 & \cellcolor{t_orange} 0.188 & \cellcolor{t_orange} 0.125 & \cellcolor{t_orange} 0.193 & \cellcolor{t_orange} 0.299 & \cellcolor{t_orange} 0.284 & \cellcolor{t_orange} 0.196 & \cellcolor{t_orange} 0.133 & \cellcolor{t_red} 0.180 & \cellcolor{t_red} 0.157 & \cellcolor{t_red} 0.243 & \cellcolor{t_red} 0.238 & \cellcolor{t_red} 0.240 \\
\bottomrule
\end{NiceTabular}}
\vspace{1em}
\caption{Per-scene results across all three datasets at 7k and 30k iterations, evaluated using structural similarity (SSIM), peak signal-to-noise ratio (PSNR), and learned perceptual image patch similarity (LPIPS).}
\label{tab:per_scene}
\end{table*} 

\section{Additional Qualitative Comparisons}
\vspace{-0.5em}

Table~\ref{tab:patches} presents additional qualitative results, demonstrating the advantages of 3DLS over existing methods. Notable improvements are observed in regions with high-frequency details and intricate textures, such as the \textbf{Bicycle}, \textbf{Counter}, \textbf{Flowers}, \textbf{Treehill}, \textbf{Train}, and \textbf{Trucks} scenes. Additionally, areas with sharp transitions, including the \textbf{Stump}, \textbf{Kitchen}, and \textbf{Dr. Johnson} scenes, exhibit reduced artifacts and improved edge preservation. These results illustrate that 3DLS effectively manages diverse textures and transitions, delivering superior visual quality in challenging regions.

\section{Additional Kernel Comparisons}

Table~\ref{tab:pm5544} presents qualitative comparisons of various kernels using the \textbf{PM5544 color pattern}\footnote{\url{https://en.wikipedia.org/wiki/Philips_circle_pattern}}, which features diverse color blocks and high- and low-frequency regions. The Linear kernel excels in high-frequency details, consistent with Section~\ref{section:kernel_distributions}, while the Cosine kernel performs best in uniform color areas, highlighting opportunities for further kernel-specific exploration.

\newcommand{\patchrow}[3]{
    \textbf{#1} & 
    \raisebox{-0.4\totalheight}{\includegraphics[width=0.12\textwidth]{figures/patches/#3_3DGS_#2.png}} &
    \raisebox{-0.4\totalheight}{\includegraphics[width=0.12\textwidth]{figures/patches/#3_2DGS_#2.png}} &
    \raisebox{-0.4\totalheight}{\includegraphics[width=0.12\textwidth]{figures/patches/#3_Mip-Splatting_#2.png}} &
    \raisebox{-0.4\totalheight}{\includegraphics[width=0.12\textwidth]{figures/patches/#3_AbsGS_#2.png}} &
    \raisebox{-0.4\totalheight}{\includegraphics[width=0.12\textwidth]{figures/patches/#3_3DLS_#2.png}} &
    \raisebox{-0.4\totalheight}{\includegraphics[width=0.12\textwidth]{figures/patches/#3_3DLS+AA_#2.png}} &
    \raisebox{-0.4\totalheight}{\includegraphics[width=0.12\textwidth]{figures/patches/#3_GT_#2.png}} \\
}

\clearpage
\begin{table*}
\centering
\setlength{\tabcolsep}{0.2em}
\renewcommand{\arraystretch}{0.35}
\resizebox{\linewidth}{!}{
\begin{NiceTabular}{@{}ccccccc|c@{}}
& \textbf{3DGS} & \textbf{2DGS} & \textbf{Mip-Splatting} & \textbf{AbsGS} & \textbf{\makecell{3DLS \\ (Ours)}} & \textbf{\makecell{3DLS+AA \\ (Ours)}} & \textbf{GT} \\
\patchrow{Bicycle}{0_100_200x200}{bicycle}\\
\patchrow{Stump}{350_0_200x200}{stump}\\
\patchrow{Counter}{1300_0_100x100}{counter}\\
\patchrow{Kitchen}{800_450_200x200}{kitchen}\\
\patchrow{Flowers}{500_400_200x200}{flowers}\\
\patchrow{Treehill}{0_350_200x200}{treehill}\\
\patchrow{Dr. Johnson}{350_100_200x200}{drjohnson}\\
\patchrow{Train}{200_0_200x200}{train}\\
\patchrow{Truck}{100_0_200x200}{truck}
\end{NiceTabular}}
\caption{Qualitative comparisons of 3DLS against existing methods on various scenes, highlighting improvements in high-frequency details, intricate textures, and sharp transitions. Examples include Bicycle, Flowers, Treehill, Train, Trucks, Stump, Kitchen, and Dr. Johnson scenes.}
\label{tab:patches}
\end{table*}

\newcommand{\imagecell}[2]{
    \raisebox{-0.4\totalheight}{%
        \ifthenelse{\equal{#2}{}}{%
            \includegraphics[width=0.15\linewidth]{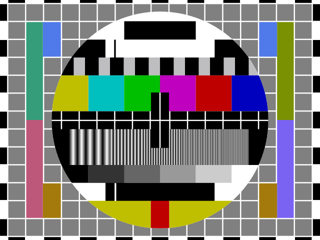}%
        }{%
            \includegraphics[width=0.15\linewidth]{figures/pm5544/#1_frame_#2.png}%
        }%
    }
}

\newcommand{\imagerow}[2]{
    \makecell{\textbf{#1}} & \foreachcategory{#2}
}

\newcommand{\gtrow}{
    \makecell{\textbf{GT}} & \foreachcategory{}
}

\newcommand{\foreachcategory}[1]{
    \imagecell{gaussian}{#1} &
    \imagecell{laplacian}{#1} &
    \imagecell{cosine}{#1} &
    \imagecell{quadratic}{#1} &
    \imagecell{linear}{#1} \\
}

\begin{table*}
\centering
\setlength{\tabcolsep}{0.05em}
\renewcommand{\arraystretch}{0.3}
\resizebox{\linewidth}{!}{
\begin{NiceTabular}{cccccc}
\makecell{} & \textbf{Gaussian} & \textbf{Laplacian} & \textbf{Cosine} & \textbf{Quadratic} & \textbf{Linear} \\ \\
\imagerow{600/3000}{599} \\
\imagerow{1200/3000}{1199} \\
\imagerow{1800/3000}{1799} \\
\imagerow{2400/3000 \quad\quad}{2399} \\
\imagerow{3000/3000}{2999} \\
\gtrow \\
\end{NiceTabular}}
\caption{Qualitative comparisons of various kernels on the PM5544 color pattern, demonstrating the Linear kernel's effectiveness in high-frequency detail regions and the Cosine kernel's superior performance in uniform color areas. The PM5544 pattern provides a mix of high- and low-frequency features and diverse color blocks for evaluating kernel behavior.}
\label{tab:pm5544}
\end{table*}

\end{document}